\documentclass[preprint,12pt]{elsarticle}

\usepackage{amssymb}
\usepackage{enumitem}

\journal{Elsevier}

\begin{document}

\begin{frontmatter}



\title{Challenges of Large Language Models for Mental Health Counseling}


\author[inst1,inf]{Neo Christopher Chung\corref{cor1}}
\cortext[cor1]{n.chung@uw.edu.pl}
\affiliation[inst1]{organization={Institute of Informatics, University of Warsaw},
            country={Poland}}

\affiliation[inf]{organization={Informatism},
            state={New Mexico},
            country={United States}}
            
\author[inf]{George Dyer}
\author[inst1,inf]{Lennart Brocki}

\begin{abstract}
The global mental health crisis is looming with a rapid increase in mental disorders, limited resources, and the social stigma of seeking treatment. As the field of artificial intelligence (AI) has witnessed significant advancements in recent years, large language models (LLMs) capable of understanding and generating human-like text may be used in supporting or providing psychological counseling. However, the application of LLMs in the mental health domain raises concerns regarding the accuracy, effectiveness, and reliability of the information provided. This paper investigates the major challenges associated with the development of LLMs for psychological counseling, including model hallucination, interpretability, bias, privacy, and clinical effectiveness. We explore potential solutions to these challenges that are practical and applicable to the current paradigm of AI. From our experience in developing and deploying LLMs for mental health, AI holds a great promise for improving mental health care, if we can carefully navigate and overcome pitfalls of LLMs. 
\end{abstract}




\end{frontmatter}


The global prevalence of mental disorders is increasing owing to a lack of treatment, services, and clinical professionals \cite{Rahman2020, Santomauro2021}. Over 658 million people suffer from psychological distress worldwide \cite{GDB2022}. In the United Kingdom, only $35\%$ of people with mental health issues receive any form of therapy or treatment \cite{Chen2021}. In this setting, the use of large language models (LLMs), recently popularized by the transformer architecture \cite{Radford2021, OpenAI2023GPT-4, Devlin2018}, presents both promising opportunities and unique challenges in the field of psychological counseling.

These AI models have the potential to assist therapists in the daily provision of mental health services, through content suggestion and patient management \cite{Stade2023, Sharma2022}. These efforts tend to focus on mental health issues that are not life-threatening and rather requires counseling. In this role, AI can help providers scale the delivery of mental health services and reduce patient costs, thus helping to address the global shortage of counselors and therapists. Additionally, several applications have been developed that place an LLM model in the role of digital counselor \cite{Serena2023, ChatCounselor2023, PsyLLM2023}. The primary challenge to all of these proposed uses is model accuracy and reliability, which are both critical for delivering ethical and effective services. In this paper, we will lay out major challenges and actionable solutions for using LLMs in the mental health field.

LLMs are a subset of artificial neural networks (ANN) that demonstrate human-like general-purpose language understanding and generation. A broad review of ML methods used in mental health counseling is available in \cite{Glaz2021}. LLMs for language generation are built on the principle of autoregression, meaning that the model is trained to predict the next token (roughly equivalent to the next word) in a given sequence of tokens. Their training is performed using large data sets of language \cite{BERT2019, BART2020, GPT32020} and the model learns to predict, given a sequence of tokens, a probability distribution over all tokens in its vocabulary for the next token in the sequence. In the current paradigm of LLMs (e.g., recurrent neural network \cite{Hochreiter1997}; transformer architecture \cite{Vaswani2017transformer}), the apparent language understanding is therefore of purely stochastic nature and amounts to predicting what's the most probable thing to ``say'' in a given context. Despite the ostensibly simple training objective of predicting the next word,'' LLMs acquire impressive capabilities when trained on extremely large sets of data \cite{CommonCrawl, ThePile}. But the models do reflect the biases and patterns existing in the training data \cite{Caliskan2017, Acerbi2023}.

Several mental health applications for use by individuals and institutions incorporate LLMs into their architecture. They can be divided into two broad categories: 1) user facing counseling and therapy; and 2) therapist assistants. Among user facing applications, we find some that provide an immersive conversation experience directly with the underlying model (e.g., \cite{Serena2023, ChatCounselor2023}), others that offer a combination of open-ended conversation with the model and rule-based elements (e.g. \cite{Woebot2022}), and finally, those that rely on the LLM primarily to understand and categorize the user’s message input, so as to better connect them with a ``real'' human therapist working for the service \cite{KokoNews2023, Sharma2022}. This last category of user facing apps may overlap with therapist assistant apps, whose generated content never directly reaches the patient. Rather, the model outputs are sent to the mental health service providers as recommendations or suggested answers, sometimes acting as a ``co-pilot.''

We have identified 5 major challenges for building, training, and deploying LLM for mental health counseling: \ref{sec:hallucination}) Model hallucination, which impacts all LLMs regardless of application; \ref{sec:interpretability}) Model interpretability, which is crucial for human understanding, wider acceptance, and model improvement; \ref{sec:ehr}) Privacy and regulatory concerns, notably arising from the use of patient electronic health records (EHR); \ref{sec:methodology}) Clinical methodology and effectiveness; and \ref{sec:bias}) Bias arising from current LLM paradigms and limited data sources.

LLMs do not possess innate empathy or have “true” understanding of conversations, although there is ongoing conceptual and practical discussion of their emergent properties \cite{Bender2020, Piantadosi2022, Zhou2023}. Even without these supposed capabilities, LLMs still can be useful for a wide range of services related to mental health. Especially, given limited resources available to treat rapidly increasing populations with mental health disorders, we must learn how to harness the many benefits of AI for therapy and counseling through a measured, risk-based approach. Our summary of challenges focuses on improving the present and the near future applications of LLMs, in order to reduce dangers and harm to patients and users. In the process, we provide potential solutions and suggestions for building and improving AI systems for mental health counseling.


\section{Model Hallucination}\label{sec:hallucination}

Hallucination is a phenomenon in which a language model generates text that appears coherent but lacks a factual grounding or accurate representation of the provided source inputs, e.g. the dialogue history (reviewed in \cite{Ji2023HallucinationReview}). All LLMs have a tendency to hallucinate, which severely limits their applicability in the medical field \cite{Hanneke2018}. This is likely a major contributing factor to the poor performance of LLMs in displaying medical knowledge and making recommendations, notably with respect to cancer \cite{Chen2023chatgpt-cancer} and liver diseases \cite{Yeo2023chatgpt-liver}. Inaccurate responses, which will likely continue to occur to some extent within any AI systems developed in the foreseeable future, are one of the biggest concerns for deploying LLMs in high risk domains like medicine. 

In the mental health area, LLMs are primarily used for dialogue generation. In this task category, there are two primary forms of hallucination: 1) Intrinsic hallucination, where the model's response is contradictory to the dialogue history or external sources of knowledge; and 2) Extrinsic hallucination, where the response cannot be readily verified against either the dialogue history or external knowledge sentences. In other words, extrinsic hallucinations are impossible to verify with the given inputs. \cite{Ji2023HallucinationReview} 
In the context of psychological counseling specifically, model hallucination can lead to several problems:

AI-generated responses may provide inaccurate or misleading information about psychological disorders, treatment options, or coping strategies, potentially leading to misguided decision-making on the part of the counselor or the individual seeking therapy. In addition, language models can inadvertently reinforce harmful beliefs or biases present in their training data 
\cite{Caliskan2017, Acerbi2023}, negatively impacting the therapeutic process. For instance, if a model suggests stigmatizing or discriminatory views toward a specific group, it can contribute to further marginalization in therapy \cite{Whittaker2019AINow}.

The experience of the user can be significantly affected when the AI generated responses demonstrate hallucinations, whether factual or simply perceived by the user. For example, if the model mis-references earlier parts of the conversation, or misstates a fact or even a belief of the user, this will negatively impact the user's trust and immersion in the session. These factors as essential to a positive experience with the service, which is a necessary condition for its therapeutic effect. Therefore, ensuring that the model produces consistent responses is essential to user acceptance, and by extension, clinical efficacy.

Model hallucination may lead to unethical and ineffective provision of mental health therapy, primarily due to the potential dissemination of inaccurate information, but also the inability to maintain an internally consistent dialogue with the user. These issues are crucial when considering the vulnerability of individuals seeking counseling and the potential consequences of inappropriate guidance. Therefore, therapists utilizing so-called "copilot" models must exercise caution and cross-reference generated responses with trusted sources. Developers of standalone models that interact directly with patients must implement effective and robust safeguards against harmful responses.

Addressing the challenges associated with model hallucination in psychological counseling requires a multi-faceted approach:

\begin{enumerate}[label=\alph*)]
\item \textbf{Data and training strategy}: A straightforward approach to improving an LLM for mental health counseling is obtaining diverse datasets. Taking into account a broader range of demographic, cultural, and clinical information can enhance the model's understanding and response generation. We also suggest incorporating domain-specific data, clinical expertise and ethical guidelines, which will enhance the LLM’s ability to generate accurate and contextually appropriate responses.  Furthermore, novel generation strategies, such as retrieval augmented generation \cite{lewis2020retrieval}, could provide better factual grounding of generated text.

\item \textbf{Safeguarding}: Likely, model hallucination will not be eliminated in the current AI paradigm. Therefore, the LLM responses should be fact-checked in evaluation phases and/or in production. We suggest a certain safeguard: for example, a hard-coded decision tree may be utilized to capture and respond to suicidal messages in a pre-determined manner. In a more advanced approach, Roy et al. used texts from abusers to create a NLP model to detect abusive texts \cite{Roy2018}. Suggestions for therapy and treatment must be checked with medical sources and known best practices.

\item \textbf{Transparent and accountable development}: Developers of language models and providers of counseling services must adopt transparent practices, openly discussing the limitations, biases, and potential risks associated with these models in therapy. Recognizing their limitations and potential for model hallucination is of utmost importance, especially in light of future proposed regulations of AI in Europe and the United States 


\item \textbf{Human-in-the-loop deployment}: Incorporating human reviewers or therapists in real time or post-hoc validation processes can help identify and rectify model hallucination instances, ensuring the provision of accurate and reliable responses. This may in some cases by a regulatory requirement, as the GDPR requires that any AI system that affects any natural person's legal rights must involve human supervision \cite{GDPR}. 
\end{enumerate}

Model hallucination presents significant challenges for the use of large language models like ChatGPT \cite{ChatGPT} in psychological counseling. Ensuring the ethical and effective implementation of these models requires a thoughtful consideration of the risks and the development of appropriate solutions. By addressing these challenges, language models can play a valuable role as an adjunct tool in psychological counseling, fostering improved access to mental health services while maintaining high-quality therapeutic practices.

\section{Interpretability of AI}\label{sec:interpretability}

The black box nature of artificial intelligence (AI) algorithms, i.e. the lack of interpretability of LLM models, poses significant challenges for their usage in psychological counseling. The most popular LLMs such as GPT are regarded as black boxes due to their complex architectures and billions of parameters. Interpreting how these models process and generate responses becomes challenging, hindering transparency and accountability \cite{Ribeiro2020}. The lack of interpretability raises concerns regarding their use in mental health counseling \cite{Sarkar2023}.

The lack of interpretability in AI models can lead to unintended biases and potentially harmful consequences in psychological counseling. Algorithms may reproduce or amplify societal biases present in the training data, resulting in biased responses, recommendations, or reinforcement of harmful narratives \cite{Mitchell2019}. It can impede a client's ability to make autonomous decisions about their therapy when AI is involved. Without understanding how decisions are reached and what information is being used, clients may feel disempowered, discriminated against, and unable to provide informed consent \cite{Kilbertus2017}.

Similarly, the black box nature of AI models can undermine the therapeutic alliance between clients and mental health professionals. Clients may perceive AI-based counseling as impersonal and question the credibility and reliability of the AI system's advice or interventions \cite{Shan2022}.

Conversely, the use of black box AI in medicine can in some instances unjustifiably increase trust. Automation bias refers to how we may perceive automated decisions, e.g., derived from AI, as more authoritative \cite{Goddard2012}. We often think of algorithms or machines as more impartial, and therefore their diagnosis must be more reliable than humans. When patients and clinicians interact with automated systems, a misdiagnosis can have a more damaging effect as they may be unwilling to question it. This automation bias would be mitigated if the AI decision making process can be clearly explained, e.g., why a certain diagnosis is made based on the patient's data.

Finally, there are regulatory pressures towards greater model explainability. For example, the EU General Data Protection Regulation (GDPR) guarantees the right to “meaningful information about the logic” of algorithmic systems, which has been interpreted as establishing “a right to explanation." Essentially, automated decisions by AI must be accompanied by a human-understandable explanation \cite{GDPR}.  

For these reasons, we foresee explainable AI (XAI) to become increasingly important in application of LLMs on mental health counseling. Essentially, XAI methods attempt to interpret why a certain response is generated by a LLM, given a prompt. Techniques such as attention visualization, feature importance analysis, and rule-based explanations can provide insights into model decision-making processes \cite{Rudin2019}. Particularly, due to the self-attention mechanisms in transformer architectures, quantifying the relationship between the prediction and the input \cite{Vig2019BertViz, Caron2021, CDAM2023} may prove to be useful for understanding the language models. 

The interpretability in LLMs must be tackled before the mainstream adaptation in psychological counseling. Ethical considerations and the need for transparency and accountability demand the development and integration of explainable AI techniques. By addressing these challenges, both human therapists and clients alike may trust AI more, ensuring patient autonomy, trust, and effective therapeutic outcomes.

\section{Electronic Health Records (EHRs)}\label{sec:ehr}

The integration of LLMs in psychological counseling can greatly benefit individuals seeking support and advice, especially by incorporating individualized data. The incorporation of patient data into the AI system, through few shot learning or prompt engineering, can vastly reduce the underlying LLM model's tendency to hallucinate. The development of "personalized" systems is the most obvious and feasible way of improving automated mental health services under the current AI paradigm. 

However, the incorporation of patient electronic health records (EHRs) into these models raises significant challenges concerning data security and patient privacy \cite{Liaw2020AI-EHR}. We explore the specific issues related to the use of EHRs with large language models in psychological counseling and discuss the potential solutions and ethical considerations necessary to safeguard patient data. Additionally, the inclusion of EHRs requires additional and careful protection of data security and patient privacy. 

The two primary regulatory frameworks that digital healthcare service providers need to work with are HIPAA (Health Insurance Portability and Accountability ACt) in the United States \cite{HIPAA}, and GDPR in Europe \cite{GDPR}. Succinctly, HIPPA is composed of two parts: a Privacy Rule and a Security Rule, both of which would come into play for any LLM-based healthcare or counseling service. For example, any organization seeking to deploy such a model would need to determine whether they would be considered a "covered entity" that deals with "designated records." If so, they would need to ensure their system meets standards such as full encryption, minimum availability, activity logging, strong passwords, employee training, emergency policies, and auditing \cite{HIPAA}. While HIPAA is focused strictly on health related services, in Europe the GDPR applies more broadly to any digital service that collects or exploits data obtained from users in the European Union. In scope data can include the user's race, religion, political affiliations, sexual preferences, biometric or genetic data, and any other information relating to their health \cite{GDPR}.

EHR can be incorporated into an AI system in two ways. First, a given user's EHR can be directly used in order to personalize their experience with the service. Through retrieval-augmented generation \cite{lewis2020retrieval}, commonly acheived through prompt engineering, the underlying LLM can be given access to the patient's data in order to provide more accurate and relevant responses. While such a set-up has the potential to provide a high quality service to the user, maintaining compliance with GDPR and/or HIPAA can be a significant challenge, both from a technical and resource perspective. Such a system raises concerns about unauthorized access to sensitive patient records, which can lead to breaches and misuse of confidential information. Strict measures need to be implemented to ensure that only authorized individuals can access and use patient data in a secure fashion. The storage and encryption of EHRs collected must meet high standards of data security. In addition, robust encryption techniques, such as end-to-end encryption with strong cryptographic algorithms, should be employed to protect patient data from unauthorized access, both during storage and transmission.

Second, EHR can be incorporated indirectly into the AI system, which may have a less noticeable effect on the user experience, but comes with lessened regulatory and ethical risk. Specifically, EHR can be used in the fine-tuning of the model, as a few shot learning resource to "point" the model towards the desired response profiles. Patient data used in large language models should be effectively de-identified and anonymized to safeguard patient privacy. Even if the records are used in an anonymous fashion, their collection and treatment up until that point will necessarily have to meet all applicable regulatory and ethical standards. Furthermore, when fine-tuned on EHR records, there is a risk of leaking patient information through model hallucinations \cite{Sarkar2023}.

In either case, transparent communication and obtaining patient consent are crucial aspects of patient privacy. Patients must be informed about the use of their EHRs in large language models and be given the opportunity to provide informed consent explicitly. Transparent policies should be in place to address patient concerns and clarify the extent to which their data will be used. Clear guidelines should be established regarding the retention and disposal of patient data used in both large language models, and the systems into which they are incorporated. Data that is no longer necessary for model improvement, or that, in the context of a system that directly uses EHR, is no longer connected to an active user, should be securely erased to prevent unnecessary retention and potential privacy breaches.

The use of patient EHR in large language models for psychological counseling holds immense promise both in improving LLMs and customizing the system's responses. EHR is already being collected and made available to the research community, which can be used to detect specific mental disorders (e.g., bipolar) \cite{Castro2015}. However, addressing the challenges related to data security and patient privacy is paramount. By implementing appropriate data protection measures, ensuring patient consent, and adhering to ethical considerations, we can harness the potential of large language models while safeguarding patient privacy in psychological counseling.

\section{Clinical methodology and effectiveness}\label{sec:methodology}

Providing emotional support within culturally and socially relevant contexts is a vital aspect of therapy. However, language models lack the capacity for genuine empathy, often resulting in responses devoid of emotional understanding and tailored guidance. Limitation of LLM’s understanding has been studied directly on users for eating disorders \cite{Chan2022eating} and depression and anxiety \cite{Lim2022-depression}. We may consider limited scopes (e.g., arising from data, architecture, training schemes, and others) inherent in the current generation of LLMs, as a key way to improve their capabilities. Furthermore, suitable counseling styles and therapeutic methodologies are necessary for beneficial AI psychological counseling.

LLMs such as ChatGPT may struggle to demonstrate emotional intelligence, resulting in inappropriate responses and an inability to understand nuanced emotional expressions \cite{Bear2022}. This limitation hampers their ability to provide sensitive and empathetic counseling. The empathic capabilities of language models are limited, and they may fail to understand the nuanced emotions and experiences shared by individuals seeking counseling. This can undermine the therapeutic alliance and hinder the provision of appropriate emotional support,

Unlike human counselors, LLMs lack the ability to process nonverbal cues and body language, which are essential for effective counseling. This deficiency limits their ability to provide appropriate nonverbal support to patients, potentially hindering the counseling process. Similarly, AI is mostly focused on texts and audio responses. And human clients may seek non-verbal cues to form authentic connections. A lack of multimodal human-AI interactions may reduce the effectiveness of the client-counselor relationship \cite{Bear2022}.

LLMs may provide misguided or inappropriate advice due to limited understanding of unique patient circumstances, leading to potential harm for vulnerable individuals. Especially, the best practices of mental health counseling are evolving with evidence-based research. For example, the current prevailing approach in mental health counseling is to be less prescriptive and advisory. But using a foundation models like GPT-4 \cite{OpenAI2023GPT-4} would often result in highly specific advice. The advice could be unwarranted and unwanted, especially since LLMs may struggle to obtain, much less understand the concept of, informed consent. As a result, patient autonomy and right to self-determination may be compromised.

The limited counseling styles of LLMs can be overcome by \textbf{a) multiple fine-tuned models, b) usage of patient data, c) real-time updates}. Particularly, there have been efficient approaches to fine-tune LLMs, such as LoRA \cite{LoRA2021} and related methods \cite{S-LoRA2023, Punica2023}. Instead of providing a single model, multiple fine-tuned models utilizing different datasets could mitigate this issue. Patient data may be utilized to effectively and seamlessly guide which model may more effectively serve their needs. Lastly, each user's chat history may be utilized to update the LLM such that counseling styles are adapted to that user's specific needs and preferences.

Given the above limitations, unprofessional counseling styles exhibited by LLMs may jeopardize patient well-being and raise ethical concerns. It is imperative to carefully evaluate and regulate their use to ensure the proper delivery of appropriate counseling styles and the protection of patients' mental health.

\section{Bias and Data}\label{sec:bias}

Many of aforementioned challenges are closely connected to the fact that the LLM’s knowledge and style are essentially tied to the training dataset. Concerns have been raised regarding the biases inherent in these LLMs \cite{Caliskan2017, Acerbi2023}, which largely stem from their training data and could negatively impact their use in psychological counseling. We explore the biases inherent in the currently utilized data, emphasizing their potential implications for psychological counseling. Biases may arise from limited training data that lack cultural and socioeconomic diversity, significantly affect the usefulness of LLMs within the context of psychological counseling. 

LLMs, including ChatGPT, are trained on vast amounts of text data obtained through web scraping from diverse sources, such as Wikipedia, PubMed Central, arXiv, Reddit, Books3, and other sources \cite{CommonCrawl, ThePile}. The trained models inherently reflect the biases present within the data sources \cite{Zhao2017, Gehman2020}, even if the datasets nowadays are in orders of terabytes – and potentially more. The training data predominantly originates from certain demographics, leading to the under-representation or misrepresentation of diverse voices and experiences. Even worse, the most recent LLMs such as OpenAI GPT-4 \cite{OpenAI2023GPT-4} and Google Bard \cite{Bard} do not reveal exactly what datasets they are trained on.

LLMs have displayed limitations in recognizing and addressing the socioeconomic status or cultural sensitivity within mental health support. These types of biases within LLMs can manifest through incorrect assumptions, stereotyping, or misinterpretation of cultural nuances when providing counseling. One of the largest data sources for LLMs is the CommonCrawl \cite{CommonCrawl, ThePile} which contains 390 TB with more than 3.1 billion pages of websites. $46\%$ of documents are in English followed by German and Russian. This focus on sources emanating from the developed "anglosphere" would inherently capture its cultural, social, and religious context. Thus, LLMs generally lack contextual understanding related to the developing world, minorities, and indigenous populations, making it challenging for them to comprehend intricate personal situations and offer tailored advice. This limitation can lead to inappropriate or ineffective support, with the potential negatively impact client outcomes. By relying on biased training data, LLMs may marginalize individuals from underrepresented groups, further excluding them from accessing appropriate and inclusive mental health resources. This exclusion further perpetuates disparities in mental healthcare access and quality. This is particularly difficult when coupled with the uni-modality of LLMs, since non-verbal communication often allow therapists to adapt to the clients need.

To mitigate biases in LLMs, developers can consider the following solutions:

\begin{enumerate}[label=\alph*)]
\item \textbf{Ethical guidelines}: LLM providers must adopt rigorous ethical guidelines, ensuring transparency, accountability, and bias mitigation efforts throughout the development process. Regular audits can help identify and rectify biases to minimize potential harm \cite{Mokander2023AuditLLM}. Unfortunately, the current trend among closed for-profit LLMs is to not reveal their data sources, let alone their pre-trained model. For example, the model and weights of Google’s medical LLMs (Med-PaLM and Med-PaLM2, a medicine-focused fine-tuned version of PaLM \cite{PaLM} and PaLM2\cite{PaLM2}) are not available despite the fact that they are published in Nature \cite{MedPaLM2023}. We urge that the application of LLMs in mental health counseling utilize open source AI models, which can be audited.

\item \textbf{Experts involvement}: The involvement of mental health professionals in fine-tuning LLMs can help mitigate biases. Collaboration between developers and mental health experts can improve the models' cultural competence and contextual understanding, and facilitate the development of inclusive and ethical guidance. One approach is to develop a LLM for mental health counseling that is specific to a culture, a language, or even a specific condition. For example, PsyLLM is a Chinese LLM for psychological counseling trained on professional $Q\&A$s and psychological articles. This may better reflect cultural norms as reflected in those Chinese datasets \cite{PsyLLM2023}.

\item \textbf{Reinforcement learning from human feedback}: Reinforcement learning from human feedback (RLHF) may be used to reduce biases in pre-trained foundation models. An extremely large dataset (typically terabytes of data) is needed to train from scratch a foundational LLM such as Google Bard \cite{Bard} or OpenAI's ChatGPT \cite{ChatGPT}. Since such a model is typically impossible for anyone but the largest corporations to train, we recommend the use of specifically designed high quality data in RLHF training which tends to have high impact on improving the accuracy and performance of the foundational model.

\item \textbf{Algorithmic faireness}: While the data plays the most important role, especially in the current AI paradigm, researchers are looking to create a deep learning model and training scheme that could reduce or eliminate algorithmic biases. For example, DeepMind explores how to achieve algorithmic fairness for LGBTQ+ communities which are often unrecorded or unobserved in the data \cite{Tomasev2021}. Particularly, there is a high prevalence of mental health problems among queer people due to high levels of stress associated with stigmatization and discrimination \cite{Mays2001, Meyer2003}. Therefore, developing LLMs for mental health counseling should take into account the experience of LGBTQ+ communities to reduce stress and to better build the therapeutic relationship.
\end{enumerate}

Mental health professionals need to be aware of the biases and limitations of LLMs \cite{Whittaker2019AINow}. Training programs should emphasize critical thinking skills and the ability to evaluate and contextualize the advice generated by LLMs, ensuring responsible and ethical use. The need for careful consideration and critical evaluation when applying LLMs in psychological counseling. By understanding and addressing the biases inherent in these models, mental health professionals can leverage their benefits while minimizing potential harm.

\section{Concluding Remarks}\label{sec:conclusion}

Numerous medical applications of LLMs have been experimented with, and to a limited extent deployed \cite{Soni2023, Lee2023, Clusmann2023}. There exist some papers demonstrating better performance of LLMs over human doctors, in terms of the patient's preference of responses \cite{Ayers2023}. On the other hand, evidence abounds of the limitations of LLMs in mental health counseling  \cite{Chan2022eating, Lim2022-depression}. Acceptance of AI in medicine, including mental health counseling, likely requires substantial improvements in multiple areas \cite{Lambert2023, Chen2022}. Both acceptance and safety requires responsible developments for psychological counseling \cite{Stade2023}.

Our five major challenges are identified and detailed from our academic research and development of consumer-facing AI services. LLMs will inevitably enter the medical field and face greater regulatory and policy pressure. Therefore, our focus is on ensuring and developing the best practices that can readily improve LLMs in the near future. Since many of these challenges are closely interlinked by overlapping concerns and solutions, we believe that a holistic approach is necessary. For example, diverse data that represents diverse populations should be collected; yet delivering fine-tuned models based on said subset of data may exacerbate any stigma associated with that population as expressed within the data. The issue is not simply one of representation, but of perspective. Furthermore, as noted in the Challenge \ref{sec:ehr}, regulatory bodies will play substantial roles likely requiring developers to demonstrate beneficial evidence of the model's use.

In the future, the current AI landscape may go through a paradigm shift where, for example, symbolic representation \cite{Kolata1982SymbolicAI, Kautz2022} and multi-modal interactions may exhibit superior performance. In the long term, one may imagine artificial general intelligence (AGI) where a true autonomous system can learn a wide range of intellectual tasks without human supervision \cite{Shevlin2019AGI, Crevier1995}. These next generations of AI hold great promise for medicine, such as eliminating model hallucination, bias, and inherent interpretability – but we cannot assume these hypothetical developments are imminent or even likely.

The prevalence of mental health disorders is growing globally, and without sufficient treatments, services, and professionals \cite{Rahman2020, Santomauro2021}, we must already begin to leverage the current AI paradigm to address the situation. Whether acting as a user-facing companion, a copilot for a counselor, or a clinical practice management tool, we must find a way to improve the resilience, relevance, and effectiveness of available LLMs in powering these tools. While the challenges are considerable, the potential benefits of using LLMs for psychological counseling are exceptional – if not necessary.

\section*{Acknowledgments}\label{sec:ack}

This research was carried out with the support of the Interdisciplinary Centre for Mathematical and Computational Modelling University of Warsaw (ICM UW) under computational allocation no. GDM-3540, the NVIDIA corporation’s GPU grant, and the Google Cloud Research Innovators program.

 \bibliographystyle{elsarticle-num} 
 \bibliography{llm-refs}





\end{document}